\documentclass[a4paper,fleqn]{cas-dc}

\usepackage[numbers]{natbib}
\usepackage{amsmath, amssymb}
\usepackage{mathtools}
\usepackage{bm}
\usepackage[ruled,vlined]{algorithm2e}
\usepackage{booktabs}
\usepackage{graphicx}
\usepackage{tabularx}

\usepackage{amsthm}                 % often already loaded by the class
\theoremstyle{definition} 

\begin{document}
\let\WriteBookmarks\relax
\def\floatpagepagefraction{1}
\def\textpagefraction{.001}
\shorttitle{EvoMorph: CFE for Continuous Time-Series Extrinsic Regression}
\shortauthors{M. Ceylan et al.}

\title [mode = title]{EvoMorph: Counterfactual Explanations for Continuous Time-Series Extrinsic Regression Applied to Photoplethysmography}                      
%\tnotemark[1,2]

%\tnotetext[1]{This document is the results of the research project funded by the National Science Foundation.}

%\tnotetext[2]{The second title footnote which is a longer text matter to fill through the whole text width and overflow into another line in the footnotes area of the first page.}

\author[1]{Mesut Ceylan}[type=editor,
                        auid=000,bioid=1,
                        orcid=0009-0006-6833-2353]
%\author[1]{Mesut Ceylan}
\cormark[1]
%\fnmark[1]
%\ead{jkk@example.in}
%\ead[url]{www.jkkrishnan.in}

\credit{Methodology, Software, Formal analysis, Investigation, Visualization, Writing – original draft}

\author[1]{Alexis Tabin}[orcid=0009-0004-4908-9250]
\credit{Writing – review \& editing}
\author[1]{Patrick Langer}[orcid=0009-0000-6669-3928]
\credit{Writing – review \& editing}

\author[1,2]{Elgar Fleisch}[orcid=0000-0002-4842-1117]
\credit{Writing – review \& editing}

\author[1]{Filipe Barata}[orcid=0000-0002-3905-2380]
\credit{Supervision, Conceptualization, Funding acquisition, Project administration, Writing – review \& editing}
%\fnmark[2]

%\ead{wjh@example.org}
%\ead[URL]{https://www.university.org}

\affiliation[1]{organization={Centre for Digital Health Interventions, ETH Zurich},
                %addressline={Jawahar Nagar}, 
                city={Zurich},
              %citysep={}, % Uncomment if no comma needed between city and postcode
               %state={Zurich},
                country={Switzerland}}

\affiliation[2]{organization={Centre for Digital Health Interventions, University of St.Gallen},
                %addressline={Jawahar Nagar}, 
                city={St.Gallen},
              %citysep={}, % Uncomment if no comma needed between city and postcode
                %state={Zurich},
                country={Switzerland}}

%\author[1,3]{T. Rafeeq}
%\cormark[2]
%\fnmark[1,3]
%\ead{t.rafeeq@example.in}
%\ead[URL]{www.campus.in}

\cortext[cor1]{Correspondence to: Weinbergstrasse 56/58, 8006 Zürich, Switzerland E-mail address: ceylanm@ethz.ch (M. Ceylan)}
%\cortext[cor2]{Principal corresponding author}
%\fntext[fn1]{E-mail address: ceylanm@ethz.ch (M. Ceylan)}
%\fntext[fn2]{Another author footnote, this is a very long footnote and
%  it should be a really long footnote. But this footnote is not yet
%  sufficiently long enough to make two lines of footnote text.}

%\nonumnote{This note has no numbers. In this work we demonstrate $a_b$
%  the formation Y\_1 of a new type of polariton on the interface
%  between a cuprous oxide slab and a polystyrene micro-sphere placed
%  on the slab.
%  }

\begin{abstract}
Wearable devices enable continuous, population-scale monitoring of physiological signals, such as photoplethysmography (PPG), creating new opportunities for data-driven clinical assessment. Time-series extrinsic regression (TSER) models increasingly leverage PPG signals to estimate clinically relevant outcomes, including heart rate, respiratory rate, and oxygen saturation. For clinical reasoning and trust, however, single point estimates alone are insufficient: clinicians must also understand whether predictions are stable under physiologically plausible variations and to what extent realistic, attainable changes in physiological signals would meaningfully alter a model's prediction. Counterfactual explanations (CFE) address these “what-if” questions, yet existing time series CFE generation methods are largely restricted to classification, overlook waveform morphology, and often produce physiologically implausible signals, limiting their applicability to continuous biomedical time series. To address these limitations, we introduce EvoMorph, a multi-objective evolutionary framework for generating physiologically plausible and diverse CFE for TSER applications. EvoMorph optimizes morphology-aware objectives defined on interpretable signal descriptors and applies transformations to preserve the waveform structure. We evaluated EvoMorph on three PPG datasets (heart rate, respiratory rate, and oxygen saturation) against a nearest-unlike-neighbor baseline. In addition, in a case study, we evaluated EvoMorph as a tool for uncertainty quantification by relating counterfactual sensitivity to bootstrap-ensemble uncertainty and data-density measures. Overall, EvoMorph enables the generation of physiologically-aware counterfactuals for continuous biomedical signals and supports uncertainty-aware interpretability, advancing trustworthy model analysis for clinical time-series applications. 
\end{abstract}

%\begin{highlights}
%\item Research highlights item 1
%\item Research highlights item 2
%\item Research highlights item 3
%\end{highlights}


\maketitle

\section{Introduction}
% Here digital health, ppg, wearables
Wearable sensing technologies have been central to digital health because they address a fundamental limitation of traditional healthcare, where physiological measurements are typically collected in episodic, clinic-bound, and infrequent snapshots \cite{Dunn2018}. In contrast, wearable technologies enable continuous and non-invasive population-scale monitoring of physiological and behavioral signals in real-life conditions \cite{Daniore2024}, capturing dynamics that are invisible to sporadic clinical encounters. Smartwatches, smart rings, and fitness trackers provide low-burden acquisition of physiological time series, allowing the detection of subtle changes in health status \cite{Shim2024}, supporting longitudinal tracking of disease risk \cite{Barata2024}, allowing early and personalized intervention. These capabilities have established multimodal wearables as a cornerstone of digital biomarker development and longitudinal health monitoring. Within this landscape, PPG is a core sensing modality that uses LEDs on the skin-surface and photodetectors to quantify changes in blood volume. Because the PPG waveform encodes the cardiovascular cycle, a single PPG waveform supports the extraction of heart rate, pulse-rate variability, respiratory rate, and surrogates of vascular properties \cite{Almarshad2022, Shin2022, Charlton2022}. 

The availability of large amounts of longitudinal, wearable, and clinical data has allowed machine learning models (ML) to estimate clinically relevant outcomes from temporal patterns \cite{Ibrahim2022, Jeanningros2024, Deng2022, Qananwah2024, Vliaho2022}. One prominent formulation in digital health is extrinsic regression, where continuous physiological signals are mapped to estimate external clinical variables rather than predict the next time point. PPG is especially well-suited for TSER tasks, offering rich opportunities to estimate outcomes such as blood pressure \cite{ElHajj2020BP}, autonomic balance \cite{Budidha2019ANS}, vascular stiffness \cite{Tanaka2011Stiffness}, or general cardiovascular state \cite{Nguyen2023}. By translating raw wearable signals into clinically actionable quantities, TSER models support risk assessment, disease monitoring, and intervention planning.

Yet predictions from black-box models are insufficient on their own: clinicians must understand why a model reached a conclusion and whether that estimate can be trusted in practice \cite{Rudin2019, Ciobanu-Caraus2024, Ennab2024}. This need has motivated a substantial body of research on explainable artificial intelligence (XAI), including feature- and attribution-based explanations \cite{Lundberg2017, Ribeiro2016}, saliency and importance maps \cite{Simonyan2014, Selvaraju2017}, and counterfactual explanations~(CFE) \cite{Wachter2017}. Among these, counterfactual explanations have gained particular attention because they align more closely with human reasoning \cite{DelSer2024TrustworthyCF}. CFE address a distinct clinical need by illustrating how the input would need to change in order to alter the model’s prediction, rather than simply marking notable parts of the signal.

% WHY CFEs matter in healthcare and what it can be used for?
However, a critical methodological gap persists: existing counterfactual methods for time series provide only limited support for generating physiologically credible counterfactual explanations for TSER. Prior work typically generates time-series counterfactuals by applying small transformations (“edits”) to the input (e.g., warping, segment replacement/deletion, or jitter) or by retrieving similar trajectories from a dataset, where similarity is commonly defined in the raw time domain using sample-wise $L_p$ distances or dynamic time warping (DTW). However, such notions of similarity do not directly encode the physiological morphology, and small time domain edits can introduce artifacts that remain "close" numerically while becoming physiologically meaningless. These methods suffer from four fundamental limitations that make them unsuitable for physiological regression tasks such as PPG-based estimation.

First, existing approaches lack explicit constraints that focus on the defining characteristics of the signal (L1). They enforce only raw time-domain closeness but overlook the key descriptors that govern waveform morphology. For example, a small phase shift can make two morphologically similar PPG waveforms appear distant under $L_2$ or $DTW$ distance, while pointwise noise-like edits can keep $L_2$ small yet introduce non-physiological spikes or spectral artifacts. Without morphology-aware objectives, modifications in the input fail to correspond to meaningful physiological changes, leaving the resulting counterfactuals partially informative at best and physiologically invalid at worst. Second, they often lack the diversity needed to reflect the range of physiologically plausible alternatives (L2). This gap is especially pronounced in physiological settings because signals exhibit substantial inter-individual and intra-individual variability, multiple plausible trajectories can yield similar clinical outcomes, and predictions often depend on temporal dynamics rather than static features. Third, their edit operators, typically segment replacements, deletion, and arbitrary warping, do not account for physiological waveform structure \cite{Delaney2021, Ates2021, Hllig2022}. These operations are prone to generate non-physiological artifacts (L3). Fourth, minimality is commonly measured using $L_p$ distances or contiguous binary masks applied directly to the raw time domain, metrics that ignore temporal alignment and frequency changes \cite{Ates2021, Wang2024}, which can be misleading for physiological signals (L4). Two waveforms may be morphologically similar (i.e., minimally altered), but misaligned in the time domain (large $L_2$), while heavily edited signals may remain deceptively close under $L_2$ despite altered frequency content or dynamics. Together, these limitations motivate a methodology that integrates physiologically grounded edit operators with morphology-aware objectives to produce diverse, coherent, and reliable counterfactuals for regression models applied to PPG waveforms and other biomedical time series modalities.

This gap matters because clinicians must understand not only why a model predicts a certain outcome, but also whether realistic and clinically achievable changes in patient's physiology could meaningfully alter the outcome. In practice, clinicians often reason exactly in these terms: for instance, asking whether a reduction in heart rate, a smoother oxygen saturation profile, or a more regular respiratory pattern would be sufficient to move a risk estimate below a treatment threshold \cite{Boland2010DecisionThresholds}. Counterfactual explanations \cite{Wachter2017} for TSER can potentially address this class of questions by constructing counter observations, feasible input signals that answer the question "how would the physiology need to differ to change the model's estimate?" Beyond interpretability, CFE enable scenario analysis, model validation, and identification of uncertain and unstable model behaviors \cite{Antoran2020, Qiu2025}, capabilities that are essential in safety-critical healthcare settings.

To address these limitations—particularly the need for multiple physiologically coherent counterfactuals—we introduce EvoMorph, a multi-objective evolutionary framework for counterfactual generation in TSER on PPG waveforms. EvoMorph combines physiologically grounded edit operators, morphology-preserving constraints, and a prescribed prediction shift to induce meaningful model-level changes while preserving clinically meaningful descriptors (e.g., amplitude, dominant frequency, plateau statistics, trend, and maximum-gradient). Using NSGA-III \cite{Deb2014}, whose reference-direction selection promotes uniform coverage of the Pareto set in objective space, EvoMorph returns a set of counterfactuals that captures distinct trade-offs between plausibility, maximum gradient regularity, and target-level change rather than a single explanation. Collectively, EvoMorph targets these limitations by shifting from raw time-domain similarity to morphology-aware constraints (L1), producing multiple feasible alternatives via a Pareto-set formulation supported by NSGA-III’s coverage mechanism (L2), using coherent edit operators (L3), and redefining “minimal change” through descriptor- and regularity-based objectives rather than $L_p$ distances or masks (L4). For evaluation, we compare EvoMorph against nearest-unlike neighbors (NUNs) \cite{Delaney2021}, i.e., real on-manifold time-series retrieved from the training set that are close to the query but differ in outcome.

Finally, we investigate whether CFE generated by EvoMorph can serve not only as plausible alternatives, but also as a mechanism for probing epistemic uncertainty in time-series extrinsic regression models. Our hypothesis is that, if the generated CFE for a given test instance remain on the local data manifold, then the variability of the model's predictions across these counterfactuals reflects the model's epistemic uncertainty. In a physiological time-series case study, we quantify uncertainty for each held-out test instance in two ways: (i) prediction dispersion over a diverse set of EvoMorph CFE generated for that instance, which defines a locally plausible counterfactual neighborhood around an unseen input, and (ii) prediction dispersion from a bootstrap ensemble evaluated on the same test instances. We then assess whether both estimators exhibit consistent uncertainty distributions across the test outcomes.

Our main contributions are as follows:
\begin{itemize}
    \item We introduce the first framework for generating various physiological CFE specifically for TSER, combining interpretable PPG morphology descriptors and plausible edit operators.
    \item We evaluate the proposed framework across three different PPG datasets against NUN baseline, assessing validity, plausibility, and diversity of the generated counterfactual sets.
    \item Through a case study on held-out instances, we demonstrate that prediction dispersion across diverse EvoMorph counterfactuals serves as a practical proxy for local epistemic uncertainty, exhibiting consistent uncertainty trends with bootstrap ensemble estimates.
\end{itemize}

\section{Related Work}
\subsection{General counterfactual explanations}
Counterfactual explanations are originally introduced as a means of articulating “minimal changes” to alter the prediction of a black-box model, typically in tabular domains with static features \cite{Wachter2017}. This line of work usually frames counterfactual search as an optimization problem over a distance measure in feature space, regularized to enforce sparsity and feasibility. Related methods discussed in the following differ mainly in how they encode feasibility, realism, and diversity.

The first group of methods focuses on a feasible and actionable recourse. FACE constructs counterfactuals by minimizing the paths in a density-weighted graph structure built over the data, so that desired changes follow paths the data distribution supports, rather than isolated points in low-density regions \cite{Poyiadzi2020}. MACE formulates counterfactual search as a constrained-optimization problem over proximity, plausibility, and user-specific actionability constraints, and highlights the need for model-agnostic recourse guarantees \cite{Karimi2019}. The second group focuses on diversity: DiCE explicitly regularizes for multiple, distinct counterfactuals that all achieve the desired prediction, proposing a framework that can produce diverse counterfactuals rather than a single optimal solution \cite{Mothilal2020}. Building on these ideas, Multi-Objective Counterfactual Explanations (MOC) approach this problem as a multi-objective problem, solved with evolutionary optimization, and returns a Pareto set of solutions that expose the trade-offs between competing criteria \cite{DandlSusanneandMolnar2020}.

\subsection{Counterfactual explanations for time series}
For time series, most work on counterfactuals has focused on classification. A prominent pattern is to view explanation as a problem of editing discriminative sequences just enough to cross the decision boundary. Native Guide follows this pattern by using a nearest-unlike neighbor (NUN) from the training data as a case-based surrogate, then interpolating between original time series and the NUN in regions identified as important by a class-activation map \cite{Delaney2021}. CoMTE also relies on NUNs, using a k-d tree and heuristic search to substitute selected subsequences of the original series with those from a target-class neighbor \cite{Ates2021}.

Temporal tweaking approaches treat the original time series as a baseline and apply a small number of local transformation via warps, scalings and offsets to alter the classifier decision, under constraints constructed to retain the transformed series close to the original \cite{Karlsson2018, Karlsson2020}. These methods are designed for classification, and the transformations are agnostic to physiological meaning: they enforce local similarity. 

More recent methods introduce structured search strategies. TSEvo uses a multi-objective evolutionary algorithm to optimize several explanation criteria while applying various time series transformations, thereby exploring a diverse set of counterfactuals \cite{Hllig2022}. Glacier formulates counterfactual generation as gradient-based search either in the original time series space or in a learned autoencoder latent space \cite{Wang2024}. Other approaches leverage discord or shapelet structure, editing subsequences that are highly discriminative to change the predicted class while keeping the rest of the series intact \cite{Bahri2024}.

Beyond classification, ForecastCF formulates CFE for time series forecasting: a forecasting model is held fixed, and gradient-based perturbations are applied to historical trajectory so that the forecasted values satisfy specified upper and lower bounds \cite{Wang2023}. This is one of the first methods to handle continuous prediction targets in a time series context. CounTS takes a different approach, defining a self-interpretable prediction model that jointly performs time series prediction and generate counterfactual trajectories in a variational Bayesian framework \cite{Yan2023}.

\subsection{Clinical counterfactuals and plausibility}
In medical AI, much of the counterfactual literature has focused on imaging. Generative approaches such as general adversarial network (GAN)-based counterfactuals aim to generate realistic alternative images that flip a model's decision \cite{Mertes2022}. Other work emphasizes spatially coherent alterations between original and counterfactual images, pointing out that clinicians expect explanations to reflect plausible disease progression or treatment effects rather than noise-like perturbations \cite{Singla2023}. Autoencoder-based counterfactuals similarly use powerful priors to discover semantically meaningful latent manifolds for both classification and regression tasks \cite{Atad2024}. Beyond per-case explanations, counterfactual scenarios have been suggested as tools for medical research, for instance to probe the effect of hypothetical changes in imaging biomarkers on downstream risk or outcome models \cite{Tanyel2023}. 

\section{Methods}
The core of EvoMorph is a multi-objective evolutionary algorithm~(Section \ref{cfeengine}), equipped with a data-centric reference set~(Section \ref{referenceset}) and PPG-specific edit operators (Sections \ref{smoothblendingcrossover} and~\ref{smoothblendingmutation}) that together define the search dynamics over candidate waveforms. The algorithm solves a multi-objective optimization problem (MOOP) whose objectives (Section~\ref{objectives}) encode physiological plausibility and diversity. The MOOP is formulated on a morphology-aware property profile (Section~\ref{counterfactualmanifold}), together with the counterfactual definition and the desired properties (Section~\ref{CFEdef}).

\subsection{Counterfactual Generation}
The input of the counterfactual generation is a trained black-box regressor $\hat{f}$ on an observation $x$ from test set. Our goal is to generate counterfactuals $x'$ which is a plausible and novel trajectory that is close to test set instance $x$ in morphology space and its prediction is in the regression target. We solve the MOOP problem with a Pareto-based evolutionary algorithm NSGA-III, adapted to optimize directly on waveforms. We initialize the search from $x$ and the reference set which we build using the Singular Spectrum Analysis (SSA)\cite{Vautard1989SSA}-denoised training set, then apply our edit operators to synthesize candidate counterfactuals. The result of the optimization is a Pareto front of candidate CFE that are scored with respect to their performance on the objectives. In the subsequent sections, we detail the core components of the generation methodology. 

\subsubsection{Objectives}\label{objectives}
The proposed MOOP consists of three objective functions that jointly enforce physiologic plausibility, artifact control, and target control. Given a trained regressor $\hat{f}$, a test instance $x$ and its ground truth label $y$, and a desired target range $\mathcal{\hat{Y}}$ (e.g., a heart-rate value), we search counterfactuals $x'$ by solving a Pareto problem:
\[
\begin{aligned}
\min_{x'} \mathcal{O}(x, x') := \quad 
& (\mathcal{O}_{\text{morph}}(x,x'), \\
& \mathcal{O}_{\text{maxgrad}}(x'), \\
& \mathcal{O}_{out}(y, \hat{f}(x'))) \\
\text{s.t}\quad\ 
 \hat f(x') \in \hat{\mathcal{Y}} 
\end{aligned}
\]

$\mathcal{O}_{morph}$ is a distance-to-morphology surrogate derived from the property profile discussed in Section \ref{counterfactualmanifold}. It enforces candidate counterfactuals to carry the morphology of the test instance while allowing the defining descriptors (e.g., amplitude, dominant frequency, plateau length) to change, if intended.

In addition, we define $\mathcal{O}_{maxgrad}$ to suppress spiky, step-like, or high-frequency artifacts by penalizing a 95th percentile of the first-difference magnitudes:
\[
\begin{aligned}
&\mathcal{O}_{\text{maxgrad}}(x') = Q_{95}(|\Delta{x'_t}|) \\
\end{aligned}
\]
where $\Delta{x'_t} = x'_t-x'_{t-1}$ are the absolute differences of the signal. 

Lastly, $\mathcal{O}_{out}$ is a margin-based error, penalizing the distance between the test set instance label, $y$, and the counterfactual label, $\hat{f}(x')$, with a margin $\delta$. It is defined as follows:

\begin{equation*}
\mathcal{O}_{out}(y, \hat{f}(x')) = \left| y - \hat{f}(x') \right| \leq \delta
\end{equation*}

\subsubsection{Counterfactual Generation Engine}\label{cfeengine}
To operationalize our multi-objective formulation, we construct a counterfactual generation engine built on NSGA-III, an evolutionary algorithm designed to maintain solution diversity in a high-dimensional objective space. The algorithm provides a mechanism for iteratively refining a population of candidate counterfactual waveforms, optimizing the balancing across the objectives. NSGA-III uses reference-point–guided selection to ensure that the optimization process preserves multiple clinically meaningful trade-offs, rather than collapsing onto a single mode of edits.

Algorithm \ref{alg:basic-opt} denotes our optimization procedure for solving MOOP to find the best counterfactual $x'$. The input of the algorithm are held-out test instance $x$, a black-box regressor $\hat{f}$, and a training set $D$. We form the population $p$ with the initialization from reference set $\mathcal{R}$, where $p=[x_1, x_2,...,x_{P}]$. Optimization starts with the initial evaluation of the population according to the three objectives. Each individual in the population receives Pareto rank, where rank-1 means non-dominated (no other solution is strictly better on all objectives). 

At each generation $g$, we perform mating selection via a tournament that prefers lower Pareto rank, then we apply crossover with probability $p_{crossover}$ (smooth blending with two parents) to exploit the well-performing individuals. We further apply mutation with probability $p_{mutation}$ (smooth blending with a randomly sampled reference individual) to explore new candidates. The resulting offsprings, $\lambda$ are evaluated and combined with the current population for environmental selection. This selection sorts and ranks the individuals based on the objectives and the resulting set becomes the next generation. This loop continues until the present number of generations $G$ is reached, yielding the best candidate counterfactuals.

\begin{algorithm}[]
\caption{Optimization Algorithm}
\label{alg:basic-opt}
\DontPrintSemicolon
\KwIn{Population size $P$, generations $G$, test instance $x$, training set $D$}
\KwOut{Best counterfactual $x'$}

$R \leftarrow \textsc{CalculateReferenceSet}(D)$\;
$p \leftarrow \textsc{InitializePopulation}(\mathcal{R})$\;
$G \leftarrow \text{maximal number of generations}$\;
\textsc{Evaluate}$(p)$\;

\For{$g \in \{0,1,\ldots,G\}$}{
    $\lambda \leftarrow \textsc{SelTournament}(p)$\;
    \For{$j \gets 1$ \KwTo $|\lambda|-1$}{
        \If{$\textsc{Random}() < p_{crossover}$}{
            $\lambda_{j-1}, \lambda_{j} \leftarrow \textsc{Crossover}(\lambda_{j-1}, \lambda_{j})$\;
        }
    }
    \For{$j \gets 0$ \KwTo $|\lambda|-1$}{
        \If{$\textsc{Random}() < p_{mutation}$}{
            $\lambda_{j} \leftarrow \textsc{Mutate}(\lambda_{j})$\;
        }
    }
    \textsc{Evaluate}$(\lambda)$\;
    $p \leftarrow \textsc{SelectNSGA}(\lambda + p)$\;
}
\end{algorithm}

%REFERENCE SET
\subsubsection{Reference Set}\label{referenceset}
The reference set is a crucial component in the optimization process to generate diverse and optimal counterfactuals. Rather than using raw time series which may contain sensor noise, motion artifacts and glitches, we construct a denoised, morphology-preserving reference set $\mathcal{R}$ from training set $D$ by applying SSA and reconstruct each time series only from its first two components: trend and oscillation. The resulting reference set is used for initialization and the smooth blending mutation operator during evolutionary optimization.

%INITIALIZATION
\subsubsection{Initialization} \label{init}
We initialize the population $p$ by sampling $P$ training instances from $\mathcal{R}$, ensuring that all initial candidates lie on the physiological manifold. Each selected instance serves as a structurally plausible starting point for constructing counterfactuals for the target test instance $x$. To enable edits during evolution, every candidate is assigned a randomly drawn window size $w_i$, sampled from the interval $[\gamma,\; 0.5T]$, where $\gamma$ is a hyperparameter that specifies the minimum editable segment length. These windows define the regions on which crossover and mutation operators act, allowing the algorithm to explore physiologically coherent variations while preserving the global signal structure.

%CROSSOVER
\subsubsection{Smooth Blending Crossover}\label{smoothblendingcrossover}
The fundamental idea of crossover is to exploit the search space during optimization by combining high-performing candidate instances to obtain diverse offsprings. However, simply cut-and-paste recombination creates discontinuities and breaks the physiological plausibility of the time series. Therefore, we propose a cosine function-based smooth blending crossover operator to produce coherent transitions across candidate individuals and preserve the morphology of the time series.   

Let $\lambda^{(1)},\lambda^{(2)}\in\mathbb{R}^{T}$ be high-performing candidate instances.
For each candidate instance $r\in\{1,2\}$, we get the window size $w_r$ and the random start index $s_r\in\{0,\ldots,\max(0,T-w_r)\}$. Then we define the blending interval:
\begin{equation*}
I_r=[\,b_r,e_r) \;=\; \big[\,\max(0,\,s_r-w_r),\ \min(T,\,s_r+w_r)\,\big]
\label{eq:blend-interval-2w}
\end{equation*}
where $b_r$, $e_r$ are the beginning and end of the blend window, respectively, and $T$ is the length of the time series. In the blending interval $I_r$, we use cosine cross-fade weights:
\begin{equation*}
\alpha^{(r)}_{u} \;=\; \beta\!\left(1+\cos\!\Big(\eta\pi + \nu\pi\,\frac{u}{|I_r|-1}\Big)\right)
\quad u=0,\ldots,|I_r|-1
\label{eq:alpha-cos-2w}
\end{equation*}
where $\beta$ is a blend factor, $\eta$ controls the phase shift and $\nu$ is the scale parameter of the cosine curve. In this case, the first offspring $\lambda_j^1$ is:
\begin{equation*}
\lambda_j^{1} =
\begin{cases}
\lambda^{(1)}_t, & t<b_1,\\[2pt]
\alpha^{(1)}_{t-b_1}\,\lambda^{(1)}_t + \big(1-\alpha^{(1)}_{t-b_1}\big)\,\lambda^{(2)}_t & t\in I_1,\\[2pt]
\lambda^{(2)}_t, & t\ge e_1,
\end{cases}
\label{eq:offspring1-2w}
\end{equation*}
where $t$ is the time index. $\lambda_j^{2} $ is defined symmetrically with $I_2$.

%MUTATION
\subsubsection{Smooth Blending Mutation}\label{smoothblendingmutation}
Unlike the crossover operator, the mutation takes advantage of the reference set $\mathcal{R}$ defined in Section \ref{referenceset} to drive optimization towards in-manifold offsprings. Similarly to the crossover operator, the mutation follows a cosine function-based blending approach to ensure coherence and omit discontinuities between individuals.

Let $\lambda\in\mathbb{R}^{T}$ be the candidate instance and $\mathcal{R}\subset\mathbb{R}^{T}$ the reference set. First, we sample reference instance $z\sim\mathrm{Unif}(\mathcal{R})$. We get the randomly assigned window size $w$ and a blending start index $s\in\{0,\ldots,\max(0,T-w)\}$. We define the blending interval as follows:
\begin{equation*}
I=[\,b,e)\;=\;\big[\,\max(0,\,s),\ \min(T,\,s+w)\,\big]
\label{eq:mut-interval}
\end{equation*}
where $b$, $e$ are the beginning and end of the blend window, respectively, and $T$ is the length of the time series. In $I$, we construct cosine cross-fade weights:
\begin{equation*}
\alpha_{u} \;=\; \beta\!\left(1+\cos\!\Big(\eta\pi + \nu\pi\,\frac{u}{|I|-1}\Big)\right)
\quad u=0,\ldots,|I|-1
\label{eq:mut-alpha}
\end{equation*}

The mutated candidate $\lambda$ is defined piecewise by:
\begin{equation*}
\lambda =
\begin{cases}
\lambda_t, & t<b,\\[2pt] \alpha_{t-b}\,\lambda_t + \big(1-\alpha_{t-b}\big)\,z_t, & t\in I,\\[2pt] \lambda_t, & t\ge e.
\end{cases}
\label{eq:mut-piecewise}
\end{equation*}

This operation locally nudges $\lambda$ toward the reference instance $z$ while maintaining continuity and preserving the signal outside $I$.

%PROPERTY FRAMEWORK
\subsection{Property Profile}\label{counterfactualmanifold}
% why naive edits fail
For physiological time series such as PPG, naive local edits are often not clinically plausible. Small pointwise perturbations can destroy waveform morphology, distort the temporal relationships, or introduce spectral artifacts that have no physiological meaning. Thus, proximity in a raw Euclidean or DTW distance alone does not guarantee compatibility with underlying cardiovascular or respiratory dynamics. Counterfactuals that satisfy the predictive target but violate these physiological constraints are clinically uninformative and can mislead interpretation.

%what do we do instead
To address this, we constrain counterfactual generation using a morphology-aware property profile i.e., a compact vector of clinically interpretable descriptors that reflects how physiology is expressed in the signal, rather than relying on representations with unclear clinical meaning. Directly perturbing individual descriptor values (e.g., values of amplitude or dominant frequency) in isolation is also undesirable because these properties are coupled in real signals and cannot be changed independently without affecting others. Instead, we measure plausibility in the joint descriptor space and penalize counterfactuals whose overall descriptor profile drifts inconsistently from the original test instance using a single integrated loss.

Accordingly, we represent each continuous time series by low-dimensional property profile vector $\phi(x)$ consisting of amplitude, dominant frequency, plateau length, baseline trend and maximum gradient. These descriptors are robust and representative, collectively controlling time-domain morphology and spectral structure. Constraining candidate counterfactuals to remain physiologically consistent in this descriptor space encourages novel yet plausible trajectories and prevents the optimizer from producing suboptimal and physiologically invalid solutions.

We define $\phi(x)$ via a small set of interpretable and physiologically plausible descriptors: 

\[
\phi(x)=\big[\phi_{\mathrm{amp}}(x),\,\phi_{\mathrm{freq}}(x),\,\phi_{\mathrm{plat}}(x),\,\phi_{\mathrm{trend}}(x),\,\phi_{\mathrm{grad}}(x)\big]
\]

The property profile vector $\phi(x)$ provides the descriptor representation used to construct the morphology-aware objective $\mathcal{O}_{\mathrm{morph}}$. In the rest of this section, we specify the components of $\phi(x)$ and show how $\mathcal{O}_{\mathrm{morph}}$  objective is formulated. 

\newtheorem*{setup}{Setup}
\begin{setup}
Let $x\in\mathbb{R}^T$ denote a PPG signal sampled at $f_s$ Hz with indices $t=0,\dots,T{-}1$, where $T$ is the signal length, and $\Delta t=1/f_s$. We set $f_s=125$.
\end{setup}

%Amplitutde
\newtheorem*{Amplitude}{Amplitude}
\begin{Amplitude}\label{def:amplitude}  The amplitude is a global scale of a signal and stable under noise and spikes. In PPG physiology, it represents the strength of the beat, particularly reflecting the magnitude of the blood volume changes. The variations in the blood volume are associated with cardiac output and indicative of the cardiovascular status. 

We exploit the robust percentile-based amplitude $\phi_{amp}(x)$ that captures the vertical span of the signal. We define $\phi_{amp}(x)$  as follows:
\begin{equation*}
\phi_{\mathrm{amp}}(x) \;=\; Q_{95}(x) - Q_{5}(x)
\label{eq:phi_amp}
\end{equation*}
where $Q_q(\cdot)$ is the $q$th percentile.
\end{Amplitude}

%Dominant Frequency
%\paragraph{Dominant frequency / heart rate.}
\newtheorem*{dominantfrequency}{Dominant frequency}
\begin{dominantfrequency} 
As quasi-periodic waveforms, PPG signals have a distinct morphology that repeats itself once per cardiac cycle. A typical PPG signal starts with the steep systolic upstroke. Following the peak of the systolic upstroke, diastolic decay occurs. During decay, the waveforms exhibit a brief upward bump, namely a dicrotic notch. The dominant frequency tracks the fundamental rhythm of the cardiac cycle by encoding these temporal characteristics.

We define the dominant frequency by forming the one–sided Discrete Fourier Transform (DFT) grid and magnitudes, then we extract the peak. The procedure is as follows:

\begin{equation*}
f_k = \frac{k}{T}f_s
\qquad k=0,1,\ldots,\left\lfloor\frac{T}{2}\right\rfloor
\label{eq:freq-grid}
\end{equation*}
where $f_k$ is the physical frequency grid and $k$ frequency-bin index. Then we compute the DFT coefficients $X[k]$, denoting the amplitude and phase of the component at $f_k$ where $x_t$ is the signal value at sample $t$.

\begin{equation*}
X[k] \;=\; \sum_{t=0}^{T-1} x_t\,e^{-i 2\pi kt/T}
\quad k=0,\ldots,\left\lfloor\frac{T}{2}\right\rfloor
\label{eq:dft}
\end{equation*}
To find the dominant bin $k^\star$, we take the magnitude of DFT coefficient $|X[k]|$ at bin at $k$ and pick the index of the largest spectral magnitude. Lastly, we map $k^\star$ back to the physical dominant frequency $\phi_{freq}(x)$.
\begin{equation*}
\phi_{\mathrm{freq}}(x) \;=\; f_{k^\star}
\qquad 
k^\star \;=\; \arg\max_{\,0\le k < \lfloor T/2\rfloor}\, |X[k]| \, 
\label{eq:dom-freq}
\end{equation*}
where $f_{k^\star} = \frac{k^\star}{T}f_s$.

\end{dominantfrequency}

%Plateau Length
\newtheorem*{plateaulength}{Plateau length}
\begin{plateaulength}
PPG pulses are asymmetric and periodic waveforms. The time spent on the systolic upstroke and the diastolic decay reflects the shape asymmetry. This characteristic is driven by the physiology of an individual (rather physiological mechanism). The duration in which the signal spends above a certain threshold encodes the time-domain contour. 

We define the plateau length descriptor as the time spent above threshold. With an adaptive threshold $\theta(x)$ (i.e., $\theta(x)=Q_{60}(x)$), we measure the fraction as follows:
\begin{equation*}
\phi_{\mathrm{plat}}(x) \;=\; \frac{1}{T}\sum_{t=0}^{T-1}\mathbf{1}\{\,x_t \ge \theta(x)\,\}
\label{eq:phi_plat}
\end{equation*}
\end{plateaulength}

%Trend
\newtheorem*{trend}{Trend}
\begin{trend}
The ordinary least squares (OLS) slope is a low-variance and phase-robust signal statistics. It represents the slow baseline component of PPG. Anchoring trend during CFE generation not only prevents non-physiologic manipulation such as shift in the baseline or slow drift, but also stabilizes other descriptors. To compute OLS slope $\phi_{trend}(x)$, we compute center time $\bar t$ and signal $\bar x$:
\begin{equation*}
\bar t=\frac{T-1}{2}
\qquad
\bar x=\frac{1}{T}\sum_{t=0}^{T-1} x_t
\label{eq:means}
\end{equation*}

Then we compute the covariance between time and signal and the variance of the time, and take the ratio to compute the slope. 
\begin{equation*}
\phi_{\mathrm{trend}}(x)
\;=\;
\frac{\sum_{t=0}^{T-1} (t-\bar t)\,(x_t-\bar x)}
     {\sum_{t=0}^{T-1} (t-\bar t)^2} \, 
\label{eq:trend}
\end{equation*}
This quantity denotes the signal units per sample, indicating how much $x$ grows through time divided by how spread out time is.

\end{trend}

% Maximum Gradient
\newtheorem*{maxgrad}{Maximum Gradient}
\begin{maxgrad}
The systolic upstroke of PPG is particularly steep, reflecting vascular dynamics. This local slope is associated with the time derivative of the blood volume. To encode the upstorke characteristic and avoid generator-driven implausible spikes, narrow pulses, strong peaks, and artifacts, we include the maximum gradient of a signal in the morphology objective. In line with other descriptors, $\phi_{grad}$ plays a crucial role in maintaining clinical plausibility during counterfactual generation.

We take the 95th percentile of absolute differences within the signal to compute the $\phi_{grad}$. 

Let $g_t=(x_t-x_{t-1})/\Delta t$ be the absolute differences for $t=1,\dots,T{-}1$, we take the 95th percentile of the $g_t$ to measure this descriptor.
\begin{equation*}
\phi_{\mathrm{grad}}(x) \;=\; Q_{95}\!\big(\,|g_t|\,\big)
\label{eq:phi_grad}
\end{equation*}
\end{maxgrad}

\newtheorem*{manfold}{Objective Formulation} \label{AVC}
\begin{manfold}
Let $x\in\mathbb{R}^T$ be the held-out test instance and $x'$ a candidate counterfactual. We define the property profile $\phi:\mathbb{R}^T\to\mathbb{R}^d$ with components $\phi(x)=(\phi_1(x),\ldots,\phi_d(x))$. For each $j\in\{1 \ldots,d\}$, we define the normalized deviation:

\begin{equation*}
\Delta_j(x,x') \;=\; \frac{\phi_j(x') - \phi_j(x)}{\,|\phi_j(x)|+\varepsilon\,}
\label{eq:delta}
\end{equation*}
with $\varepsilon>0$ to avoid division by zero. For each property $j$, we specify a mode $m_j\in\{\mathrm{preserve},\mathrm{change}\}$ and a weight $w_j>0$. We penalize any deviation for ``preserve'' properties, and enforce a minimum relative change $\tau_j>0$ for ``change'' properties via a hinge:

\begin{gather*}
\mathcal{O}_{\mathrm{morph}}(x,x')
\;=\; \sum_{j} w_j \,\ell_j\!\big(\Delta_j(x,x')\big) \\
\qquad \ell_j(\Delta)=
\begin{cases}
|\Delta|, & m_j=\mathrm{preserve}\\[4pt]
\max\{0,\,\tau_j - |\Delta|\}, & m_j=\mathrm{change}
\end{cases}
\label{eq:manloss}
\end{gather*}

Intuitively, $\mathcal{O}_{\mathrm{morph}}$ is zero when all preserve-properties are unchanged and all change-properties have moved by at least $\tau_j$. 

\end{manfold}

\subsection{Formal Definition of Counterfactuals in TSER} \label{CFEdef}
In this section, we formalize CFE for TSER to ground the objective design introduced in Section~\ref{objectives}. The same definition also motivates the TSER-specific desired properties presented next, from which we derive the performance assessment metrics used in our evaluation (Section~\ref{performancemetrics}).
% CFE DEFINITION
Let $\hat{f}: \mathcal{X} \rightarrow \mathcal{Y}$ be a trained black-box regressor and let $\mathcal{D}$ denote the underlying data distribution over pairs $(x,y)$, where $x\in\mathbb{R}^{T}$ is a univariate time series and $y\in\mathbb{R}$ is a continuous extrinsic target.
In our experiments, counterfactuals are generated for query instances $x$ drawn from a held-out test set; however, the definition below applies to any query $x\sim\mathcal{D}$.

Given a query time series $x$ and a user-specified desired outcome set $\hat{\mathcal{Y}}\subseteq\mathbb{R}$, a counterfactual explanation $x'\in\mathbb{R}^{T}$ is an alternative time series that satisfies:

\begin{itemize}
    \item \textbf{Validity:} its prediction is in the desired outcome set i.e., $\hat{f}(x') \in \hat{\mathcal{Y}}$.
    \item \textbf{Morphological proximity:} $x'$ is morphologically proximal to $x$ and carries its characteristics.
    \item \textbf{Sparse editing:} $x'$ differs from $x$ through limited spectral modification while representing a non-trivial alternative trajectory.
    \item \textbf{On-distribution plausibility:} $x'$ is consistent with the data distribution, i.e., $x' \sim \mathcal{D}$.
\end{itemize}

%DESIRED PROPERTIES
\subsubsection{Desired Properties of CFE in TSER Problem}\label{desiredprops}
Building on standard counterfactual desiderata from prior work, we formalize TSER-specific desired properties for continuous time-series counterfactuals. These properties serve as quality measurement for each CFE instance. We define the properties as follows:
%VALIDITY
\newtheorem*{validity}{Validity}
\begin{validity}
Because extrinsic regression predicts a continuous quantity (e.g., heart rate or respiratory rate) rather than a class label, a counterfactual is only useful if it answers a specific what-if question about changing that outcome. Accordingly, we define the desired outcome set $\hat{\mathcal{Y}}$ as a range that is $\hat{\mathcal{Y}}=[\hat{y}_{min}, \hat{y}_{max}]$. We instantiate the desired outcome set per test instance as a symmetric interval around its ground-truth label $y$: $\hat{\mathcal{Y}} = [\hat{y}_{\min}, \hat{y}_{\max}]=[y - \delta,\; y + \delta]$ where $\delta > 0$ controls the tolerated deviation from the target outcome. The counterfactual explanation $x'$ is valid only if it fulfills the following constraint: $\hat{f}(x') \in \hat{\mathcal{Y}}$.
\end{validity}

%PROXIMITY
\newtheorem*{proximity}{Proximity}
\begin{proximity}
The core idea of the proximity property is to keep the counterfactual $x'$ near the data manifold $D$. If allowed, the large edits and deformations would push counterfactual off-manifold, towards sensor artifacts, non-physiologic and discontinuous forms. Therefore, we keep the counterfactual $x'$ close to its corresponding test instance $x$, ensuring morphologically minimal changes and credible counterfactual. We measure the normalized proximity between $x'$ and $x$ with DTW as follows: $d_{\mathrm{DTW}}(x,x') =\ \frac{1}{T}\,\mathrm{DTW}(x,x')$ where $T$ is the length of $x$. 
\end{proximity}

\newtheorem*{plausibility}{Plausibility}
\begin{plausibility}
Unlike proximity where the closeness between test instance and counterfactual is measured, the plausibility measures the label consistency in the local neighborhood of the counterfactual. Plausible CFE should reside in the regions of the training manifold where similar waveforms retain similar labels. Using k-nearest neighbors (k-NN) search in time-series space with DTW similarity distance, we retrieve $k$ closest training signals. Plausibility is the fraction of neighbors whose labels fall within a meaningful tolerance $\delta$ of the counterfactual's predicted value. Formally, we define plausibility as follows:
\begin{gather*}
P(x') = \frac{1}{k}\sum_{i \in \mathcal{N}_k(x')} \mathbf{1} \{|y_i - \hat{y}| \le \delta\}
\label{eq:p}
\end{gather*}
where $\mathcal{N}_k(x')$ is the index set retrieved from k-NN for corresponding counterfactual $x'$.

\end{plausibility}

\newtheorem*{tempspars}{Time Domain Temporal Sparsity}
\begin{tempspars}
The local perturbations on quasi-periodic biosignals, such as PPG, risk producing artifact-like and physiologically unrealistic waveforms. Many unique biosignals share the same morphology and temporal structure at the level of clinically reasonable descriptors such as amplitude range, plateau behavior, and slope statistics. Therefore, we require morphology-level proximity through sparse, localized edits and a novel trajectory in our counterfactuals instead of equality of signal samples. We measure the temporal sparsity $x$ and $x'$ with ${\text{S}(x, x')}$ where $S$ is the change segment count.

Let $d_l \in \{0,1\}$ be a binary difference mask:
\begin{equation*}
d_l =
\begin{cases}
1, & \text{if } x_l \neq x_l'\\
0, & \text{otherwise}
\end{cases}
\qquad l=1,\dots,L
\end{equation*}
and define $d_0=0$ and $\Delta d_l = d_l - d_{l-1}$. The change-segment count is:
\begin{equation*}
S(x,x') \;=\; \sum_{l=1}^{L} \mathbf{1}\{\Delta d_l = 1\}
\end{equation*}
which counts the number of edited segments (transitions $0\!\rightarrow\!1$) in the difference mask.

\end{tempspars}

% FREQUENCY DOMAIN SPARSITY
\newtheorem*{freqspars}{Frequency Domain Sparsity}
\begin{freqspars}
When defining counterfactuals in TSER problems, considering only the time domain of the time series is incomplete. The unique characteristics of the signal are fundamentally captured by its spectral signature. Therefore, a plausible and realistic counterfactual is a new trajectory that preserves the spectrum characteristics of the test instance, except that the change is intended. To measure morphologic fidelity of the counterfactual and its closeness to the spectral manifold, we propose frequency domain sparsity principle that quantifies the difference by by calculating Kullback–Leibler (KL) Divergence of the magnitude spectrum of the observation and the counterfactual.

For each time series $x$ and $x'$, we compute the one-sided Fast Fourier Transform (FFT) magnitudes and group them into $B$ predefined magnitude intervals (bins). This yields two empirical histograms over magnitude levels. We then normalize these histograms (with a small $\varepsilon$ added to avoid zero probabilities) to obtain discrete probability distributions $h_x$ and $h_{x'}$ over magnitude levels.
The frequency–domain sparsity score is the KL divergence between these two distributions:
\begin{equation*}
D_{\mathrm{KL}}\!\big(h_x \,\|\, h_{x'}\big)
\;=\;
\sum_{j=1}^B h_x(j)\,\log\!\frac{h_x(j)}{h_{x'}(j)}
\end{equation*}

A value of zero indicates identical magnitude–level distributions; larger values indicate the increasing discrepancy in the spread of spectral magnitudes. Note that this measure compares distributions over magnitude levels rather than frequency bins.

\end{freqspars}

\subsection{Case study: uncertainty quantification via CFE}\label{uncertainty}
%NEW
As an additional and stringent validation of counterfactual fidelity, we test whether generated CFE can be used to approximate epistemic uncertainty on unseen data. For each held-out BIDMCHR test instance, we compare prediction dispersion from a bootstrap ensemble with prediction dispersion across EvoMorph-generated counterfactuals for that instance.
 
As a reference, we train an ensemble of $B\!=\!20$ bootstrapped replicates of the regressor $\hat{f}$ by resampling the training sets with replacement. For each test sample, we obtain set of predictions, from which we compute the empirical mean, a central 95$\%$ confidence interval, and its width as a measure of ensemble spread. In addition, we fit gaussian kernel density estimator (KDE) on the training data and evaluate the negative log-likelihood (NLL) of each test sample, using NLL as a density-based proxy for how well each test sample is supported by the training distribution. 

For each test sample, we generate diverse CFE using EvoMorph and then apply the same $B\!=\!20$ bootstrap regressor to all CFE and compute, for each test sample, the mean counterfactual prediction, a central $95\%$ interval over the counterfactual predictions, and the variance of these predictions as a heuristic measure of local epistemic uncertainty around test instance. Finally, we bin test samples by ground-truth target range and, within each bin, summarize the means of bootstrap and counterfactual interval widths, KDE NLL, and counterfactual variance. 

In Section~\ref{casestudyres}, we report how the CFE-based proxy relates to ensemble spread and data density, and how all three jointly characterize regions of increased model uncertainty.

\section{Experiments}
% DATASETS
\subsection{Datasets}\label{dataset}
We use three publicly available time series datasets from Monash, UEA, and UCR Time Series Extrinsic Regression Repository, particularly BIDMCHR, BIDMCRR, BIDMCSpO2 datasets \cite{Tan2020TSER}. All datasets are extracted from the much larger MIMIC II waveform database and consist of PPG and ECG recordings from 53 patients. In this work, we focus on the PPG dimension to generate CFE. We use BIMDCHR and BIDMCRR datasets to regress heart and respiratory rates respectively. To estimate oxygen saturation, we use the BIDMCSpO2 dataset. During our experiments, we use the predefined training and test set splits to train and evaluate the Inception Network (see Section~\ref{inception}). Table \ref{tab:dataset_stats} demonstrates the dataset statistics.

\begin{table}[!htbp]%[width=\linewidth]
\caption{Dataset statistics of BIDMCHR, BIDMCRR, and BIDMCSpO2 datasets.}
\centering
\resizebox{\columnwidth}{!}{%
\begin{tabular}{@{}lcccccc@{}}
\toprule
\textbf{Dataset } & \textbf{Training set} & \textbf{Test set} & \textbf{Signal length} & \textbf{Age} & \textbf{Sex} \\ 
\midrule
BIDMCHR   & 5550 & 2399 & 4000 & \multirow{3}{*}{63.3±16.4} & \multirow{3}{*}{{51\% female}} \\
BIDMCRR   & 5471 & 2399 & 4000 &                                       &                                      \\
BIDMCSpO2 & 5550 & 2399 & 4000 &                                       &                                      \\
\bottomrule
\end{tabular}%
}
\label{tab:dataset_stats}
\end{table}

%NEAREST UNLIKE NEIGHBORS
\subsection{Baseline}
We use NUNs as our baseline counterfactuals for two main reasons. First, NUNs are real time-series instances drawn directly from the training set \cite{Delaney2021}; thus, they lie on the empirical data manifold and naturally preserve clinically plausible physiological structure. Second, because NUNs are selected to be close to the observation while differing in their associated outcome, they serve as counterfactual surrogates that reflect realistic temporal trajectories without requiring synthetic edits.

Most of the existing counterfactual generation methods are designed for time series target classification tasks and rely on transformation mechanisms that minimally modify segments of the original observation (e.g., \cite{Delaney2021, Hllig2022, Wang2024, Karlsson2020}. Such segment-level perturbations are ill-suited to our regression setting, as small localized edits can lead to physiologically implausible or internally inconsistent signals. In contrast, NUNs provide a principled and clinically reliable comparison baseline because they are (i) close to the query instance in the input space, and (ii) drawn directly from the data distribution, ensuring that their full trajectories remain physiologically valid.

\subsection{Black Box Regressor}\label{inception}
 
For the regression model, we use the Inception Network, an Inception-style convolutional architecture adapted for time-series regression \cite{IsmailFawaz2020}. Its multi-scale convolutional design captures patterns at different temporal resolutions, making it well suited for PPG-based TSER. We train and evaluate the model using three univariate PPG datasets indicated in Section \ref{dataset}. The exact network architecture and training methodology are explained in Appendix \ref{appendixmlarc}.    

\subsection{Experimental Setup}
We conduct experiments in two stages. First, we train and tune separate black-box regressor for each dataset. Second, we generate EvoMorph counterfactual explanations using NSGA-III with the pretrained models.
\newtheorem*{step1}{Step 1: Train the regressor and build the reference set}
\begin{step1}
We train the black box regressor model $\hat{f}$ on the official training set and evaluate on the held-out test set. From the training set, we reserve 20\% as a validation set for hyperparameter tuning. After model selection, we save the checkpoint with the best validation performance for use during counterfactual generation. 

Because counterfactual generation requires repeatedly querying $\hat{f}$ during optimization, we keep the trained model fixed (inference mode) and treat it as a black-box regressor. Model parameters are not updated; predictions from $\hat{f}$ define the task-specific objective(s) in our multi-objective formulation.

Before the optimization step, we construct a reference set derived from the training data to support physiologically informed edit operators. Prior to inclusion, all signals undergo SSA denoising (Section \ref{referenceset}) to remove high-frequency artifacts while preserving physiologically meaningful components such as dominant oscillatory modes and baseline trends. This reference set provides the building blocks for our crossover and mutation operators in Step 2, ensuring that population initialization and subsequent edits remain aligned with the intrinsic morphology of the underlying physiological manifold.
\end{step1}

\begin{figure*}[!htbp]
	\centering
	\includegraphics[width=\textwidth]{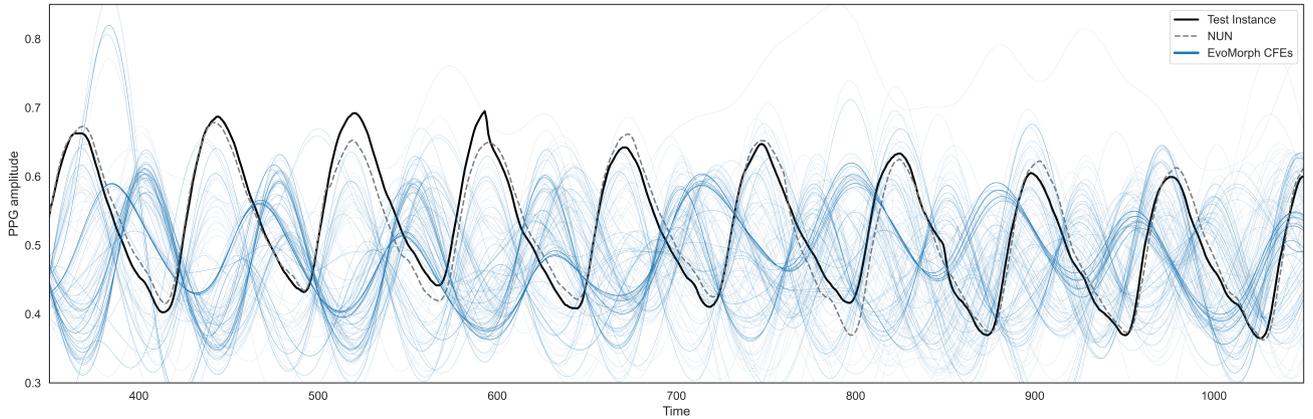}
	\caption{Waveform visualization (zoomed to samples 350-1200 within 4000-sample window) of a held-out BIDMCHR test instance (black), its NUN (gray dashed), and the full set of 251 diverse EvoMorph CFE (blue) generated for that test instance.}
	\label{diversityfig}
\end{figure*}
\newtheorem*{step2}{Step 2: Generate counterfactuals with NSGA-III}
\begin{step2}
To search for plausible counterfactual time series that would change the prediction of the trained regressor $\hat{f}$, we apply the multi-objective evolutionary algorithm, NSGA-III. NSGA-III is well-suited to this problem because counterfactual generation in TSER involves optimizing multiple, potentially conflicting objectives related to predictive validity, signal coherence, and morphological similarity.

Each candidate counterfactual $\lambda$ represents a full time-series instance of the same length and sampling rate as the held-out test instance $x$. Rather than constructing the initial population by editing the specific instance $x$, we initialize the population directly from the reference set to ensure that all individuals start from physiologically plausible signals. For each individual, we additionally sample a set of window configurations (i.e., start and end indices defining local segments) at random. These windows specify the segments that will be subject to crossover and mutation during the evolutionary process.

At each generation, NSGA-III evaluates all individuals (parents and offspring) using the three objective functions described in Section \ref{objectives}. During optimization, the trained regressor $\hat{f}$ is kept fixed and queried repeatedly to evaluate the predictive objective and to determine whether each candidate satisfies the desired outcome constraint. Candidates that meet the predictive constraint are treated as feasible and ranked based on the objectives mentioned above, while infeasible candidates are penalized via $\mathcal{O}_{morph}$ to push them toward the physiological data manifold.

Crossover and mutation are implemented using TSER-specific operators introduced in Sections \ref{smoothblendingcrossover} and \ref{smoothblendingmutation}, each acting on the predefined windows assigned to every candidate waveform. Crossover exchanges coherent subsequences between parent time series within their respective window locations, preserving local temporal structure. Mutation replaces the windowed segment of a candidate with the corresponding segment from a randomly selected reference-series instance, enabling controlled exploration of physiologically plausible variations. By restricting all edits to these predefined windows, the operators enforce locality within the morphological space and promote coherent, interpretable blending of signal characteristics.

Across $G$ generations, NSGA-III maintains a diverse set of candidate solutions via reference-point-based niching. The diversity is essential to allow the algorithm to explore multiple causal pathways and counterfactual mechanisms (e.g., amplitude-mediated vs. frequency-mediated adjustments) that could lead to the desired outcome. After optimization, we extract the final non-dominated set as the set of Pareto-optimal counterfactuals, each representing a distinct, physiologically plausible modification that changes the model’s prediction in a desired direction. 

Figure \ref{diversityfig} demonstrates over two hundred diverse CFE generated for single heart rate test instance.  Figure \ref{nsgahypesandconv} in Appendix \ref{objmoe} summarizes the evolution of the multi-objective optimization objectives across generations for each dataset and reports key performance indicators of the MOOP process, including diversity, hypervolume, and convergence.
\end{step2}

\subsection{Hyperparameters of CFE generation }\label{cfehyperparameters}
We use a fixed configuration of hyperparameters across all datasets to ensure comparability. The settings can be grouped into three categories: (i) NSGA-III optimization parameters controlling the search optimization, (ii) evaluation tolerances for the Validity and Plausibility metrics, (iii) the configuration of the morphology-based objective, specifying which descriptors are preserved or allowed to change and how strongly they are weighted, and (iv) blending parameters of the edit operators. Table \ref{hyperparametertable} in Appendix \ref{appendixhype} summarizes these hyperparameters.

\subsection{Performance Metrics}\label{performancemetrics}
In this study, we have two sets of evaluation metrics, one for the regression performance and one for the quality of the generated counterfactuals. 

%Besides the model performance, we evaluated the 
\subsubsection{Counterfactual Performance Metrics}\label{cfeperformancemetrics}
We evaluate counterfactual quality using the set of desired properties defined in Section \ref{desiredprops}, which operationalize core requirements for clinically meaningful CFE, including outcome validity, physiological plausibility, and minimality of change.

Validity quantifies the fraction of generated counterfactuals whose predicted outcome satisfies the desired target condition. In the context of extrinsic regression, we define this range as a symmetric interval of width $2\delta$ around the ground truth label of the test instance. Validity therefore captures the functional success of the counterfactual generation process: higher values indicate a larger proportion of CFE achieve the target output.

Plausibility evaluates the local neighborhood of the counterfactual by comparing the label values using k-NN. Considering a counterfactual can be numerically close to the query instance, yet implausible due to edits, this metric measures the counterfactual's fidelity to the data manifold. In our experiment, we set k=5 to report the performance of this metric. The higher values indicate that counterfactual is in a densely populated and label-consistent region, whereas the lower values denote counterfactual is off-manifold. Therefore, the higher value of this metric is ideal.

Proximity is a DTW-based normalized distance metric measuring the temporal similarity between the counterfactual and the observation. Lower values denote that counterfactual is structurally closer to the observation, which is preferable. 

Temporal and Frequency sparsity measure the extend of modifications introduced by the generation process. Temporal sparsity quantifies the number of altered time-domain segments, whereas frequency sparsity computes the divergence between spectral magnitude distributions using KL divergence. Lower values of Frequency sparsity indicate more localized and morphologically minimal edits, which are desirable for interpretability and physiological coherence.

\begin{table*}[!htbp]
%\centering
%\resizebox{\columnwidth}{!}{%
\begin{tabular}{c|cc|cc|cc}
\hline
\multicolumn{1}{c|}{Dataset} & \multicolumn{2}{c|}{BIDMCHR} & \multicolumn{2}{c|}{BIDMCRR} & \multicolumn{2}{c}{BIDMCSp02} \\ \hline
Performance Metric& EvoMorph  & NUN  & EvoMorph & NUN & EvoMorph & NUN \\ \hline
Validity $\uparrow$         & \textbf{0.929} & 0.749 & 0.903 & \textbf{0.983} & 0.919 & \textbf{0.968}  \\    
Plausibility $\uparrow$    & 0.405 & \textbf{0.563} & \textbf{1.0} & \textbf{1.0} & \textbf{1.0} & 0.999  \\    
Proximity $\downarrow$       & 0.003 & \textbf{0.002} & 0.001 & \textbf{0.001} & 0.001 & 0.001  \\
Temporal Sparsity          & 1.0 & 0.011 & 1.0 &  0.006 & 1.0 & 0.007 \\
Frequency Sparsity $\downarrow$ & \textbf{0.009} & 0.011 & \textbf{0.007} & 0.010 & \textbf{0.009} & 0.010    \\
Maximum Gradient $\downarrow$  & 0.238 & \textbf{0.074} & 0.166 & \textbf{0.074} & 0.185 & \textbf{0.072}     \\
Diversity $\uparrow$         & \textbf{247.2}   & 0  & \textbf{508.3} &  0   &  \textbf{458.1}    &  0   \\ \hline
\end{tabular}
%}
\caption{Quantitative results table, depicting the performance comparison of EvoMorph with baseline NUN across datasets and counterfactual performance metrics. $\uparrow$ indicates the higher value is better, whereas $\downarrow$ indicates the lower is better across these metrics.}
\label{resultstable}
\end{table*}

Maximum gradient evaluates overall signal smoothness by measuring the largest pointwise gradient across the time series. Lower values correspond to counterfactuals free from random peaks, spikes and artifacts. 

Diversity measures the number of distinct, plausible counterfactuals generated per test instance. Higher diversity indicates that the method identifies multiple alternative physiological pathways capable of producing the desired outcome.

\subsubsection{Machine Learning Performance Metrics}\label{mlperformancemetrics}
We evaluate our black box regressor model with the established performance metrics for regression task: mean absolute error (MAE) and mean squared error (MSE).

\section{Results}
\subsection{Quantitative Evaluation of CFE}
We compare EvoMorph with baseline NUN based on seven performance metrics introduced in Section \ref{cfeperformancemetrics}.  

Table \ref{resultstable} summarizes the quantitative evaluation across three PPG datasets. For Validity, EvoMorph reliably produces outcome-compliant counterfactuals in BIDMCHR (92.9\% vs. 74.9\%). However, NUNs achieve higher Validity in BIDMCRR and BIDMCSpO2 (98.3\% and 96.8\%). Plausibility reveals the complementary strengths of the two approaches. NUNs show higher Plausibility in BIDMCHR (56.3\% vs. 40.5\%) while both methods achieve high and comparable performance in BIDMCRR and BIDMCSpO2 datasets. For Proximity, both methods remain very close to original signals across all datasets, with DTW distances on the order of 0.001-0.002. NUNs are slightly closer for BIDMCHR dataset, while the distances are effectively identical for the other two datasets. The Temporal Sparsity metric is one across datasets for EvoMorph as this framework produces a complete new signal. Frequency Sparsity metrics consistently favor EvoMorph, which achieve lower Temporal Sparsity than NUNs. The Maximum Gradient metric is lower for NUNs in all datasets (0.074-0.072), whereas EvoMorph incurs higher gradients across datasets compared to NUNs (0.238, 0.166, and 0.185). Finally, Diversity indicates the capabilities of EvoMorph on producing alternative instances: while NUNs provide only a single surrogate per observation by design (therefore Diversity is zero in all cases), EvoMorph generates hundreds of distinct counterfactuals. 

\subsection{Case study results: uncertainty assessment with CFE diversity}\label{casestudyres}

\begin{figure*}[!htbp]
	\centering
	\includegraphics[width=\textwidth]{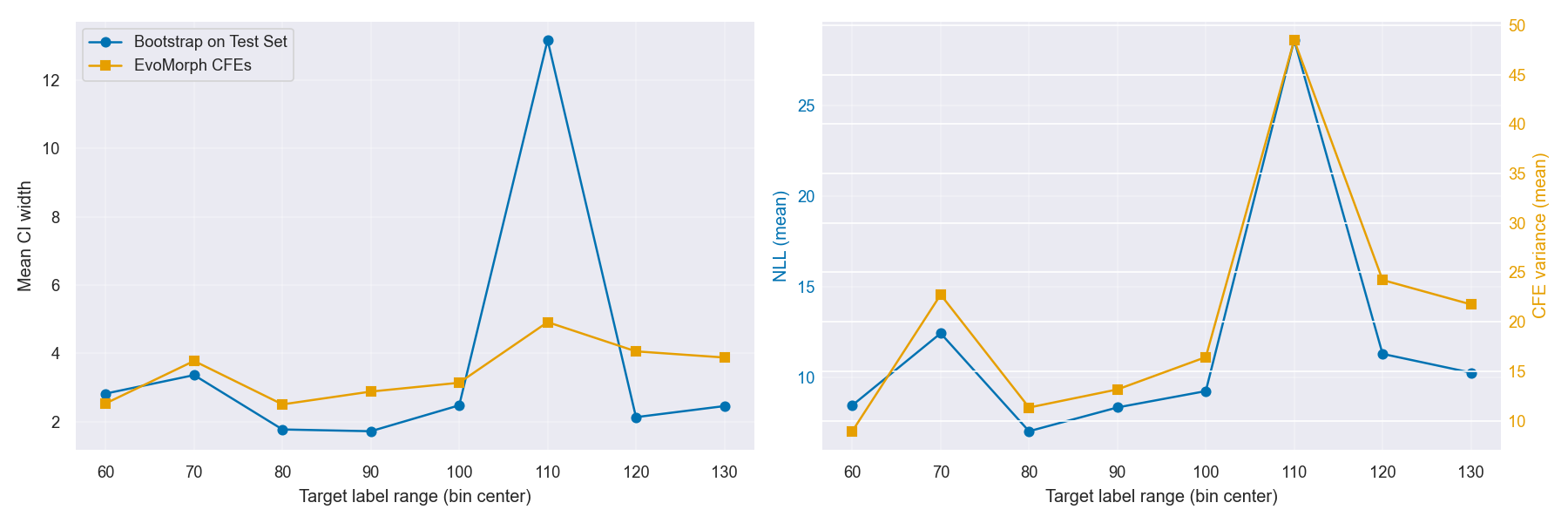}
	\caption{Epistemic uncertainty and data density across target-value bins. Left: Mean confidence interval width as a function of the target-value bin center, comparing bootstrap-ensemble uncertainty computed from predictions on the test set with prediction dispersion over the set of  EvoMorph CFE generated for each test instance. Right: Mean KDE NLL (inverse proxy for data density) and mean CFE-induced prediction variance across target bins, illustrating the correspondence between low data support and increased epistemic uncertainty.}
	\label{cfesuncertainty}
\end{figure*}

We investigate the relationship between bootstrapping-based uncertainty, counterfactual-based uncertainty, and data density by binning the test samples according to their ground-truth heart rate and summarizing uncertainty metrics within each bin. Figure \ref{cfesuncertainty} depicts the outcomes of the uncertainty assessment.

The case study results show that bootstrap-based prediction intervals are relatively narrow and stable across most bins (1.7–3.4 units), but exhibit a marked widening in [105,115) (13.17), indicating substantially increased uncertainty in this high-target region. CFE-based intervals are generally slightly wider than bootstrap intervals at lower and mid-range targets, with a moderate peak in [105,115)
(4.91), but remain more homogeneous across bins. KDE-based NLL and mean CFE variance show a similar pattern, with pronounced maxima in 
[105,115) range (28.61 and 48.49, respectively), consistent with reduced data density and elevated epistemic variability in that range, while remaining comparatively moderate in the other bins. Table \ref{tab:epistemictable} in Appendix \ref{appbepistemic} provides the further details of the epistemic uncertainty case study results.

\subsection{ML Performance on TSER}
Across three datasets, the Inception Network achieved MAE values of 4.49-5.60 and MSE values 32.0-43.5. Table \ref{tab:mlperformance} demonstrates the dataset-wise performances.

\begin{table}[!htbp]
\caption{The performance of Inception Network.}
\centering
%\resizebox{\columnwidth}{!}{%
\begin{tabular}{lcc}
\toprule
\textbf{Dataset } & \textbf{MAE} & \textbf{MSE} \\ 
\midrule
BIDMCHR   & 4.49 & 32.04  \\
BIDMCRR   & 5.60 & 43.47 \\
BIDMCSpO2 & 5.41 & 41.87 \\
\bottomrule
\end{tabular}%
%}
\label{tab:mlperformance}
\end{table}

\section{Discussion}
%MAIN FINDINDS
In this study, we introduce EvoMorph, a morphology-aware framework for generating CFE tailored for TSER problems on PPG signals, by combining interpretable descriptors, plausible edit operators, and an evolutionary algorithm. 

\begin{figure*}[!htbp]
	\centering
	\includegraphics[width=\textwidth]{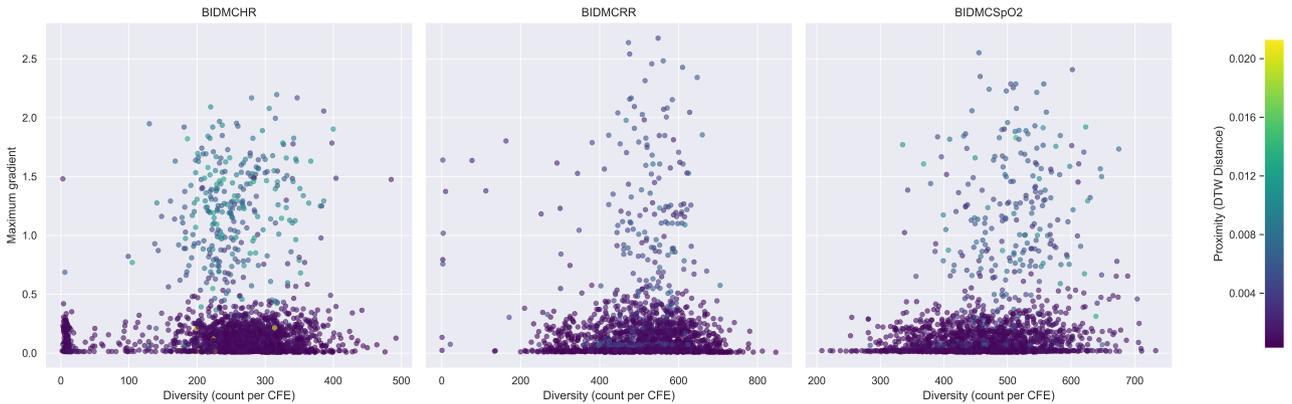}
	\caption{Diversity–smoothness–proximity trade-offs of EvoMorph counterfactuals across three PPG TSER tasks. Each point represents EvoMorph CFE; the x-axis reports per test instance number of diverse CFE generated, and the y-axis reports the maximum-gradient metric (a proxy for local waveform sharpness/spike-like artifacts). Point color encodes DTW-based proximity to the held-out test instance (darker = closer).}
	\label{divproxval}
\end{figure*}

\begin{figure*}[!htbp]
	\centering
	\includegraphics[width=\textwidth]{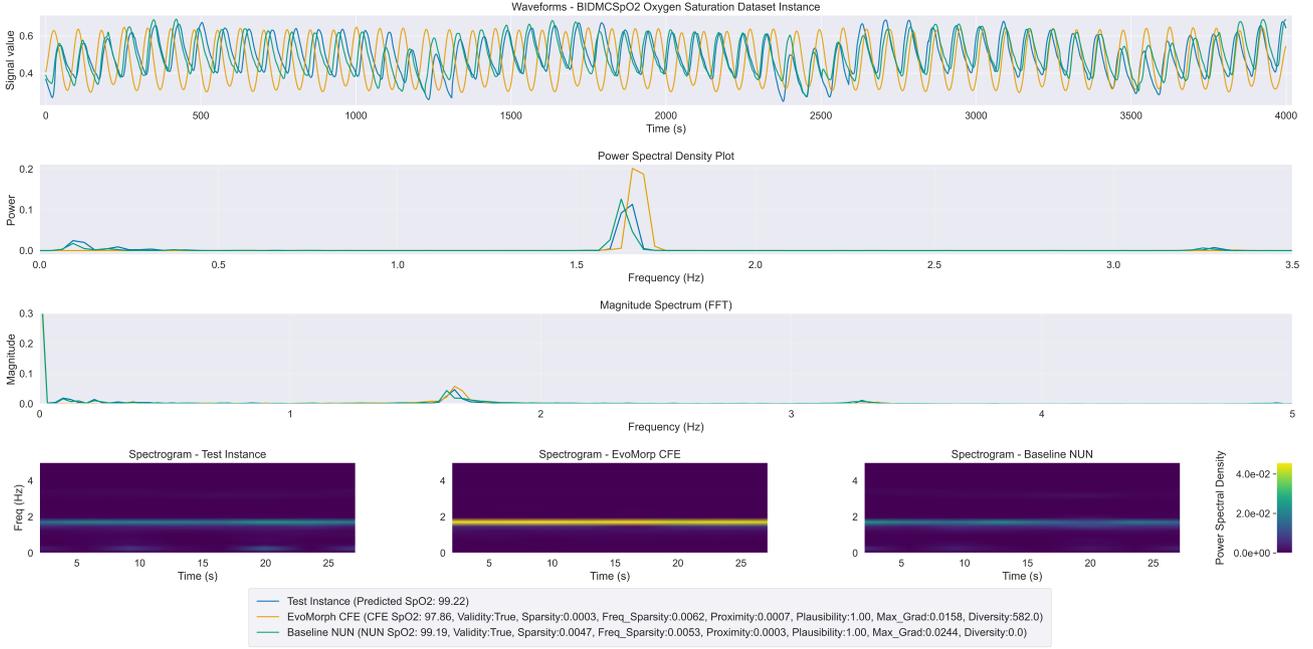}
	\caption{Qualitative comparison of the held-out test instance (blue) from BIDMCSpO2 (oxygen saturation dataset), its NUN (green), and EvoMorph counterfactual for a representative oxygen saturation signal (yellow). Top to bottom: (1) Time-domain waveforms; (2) Power spectral density plot; (3) Magnitude spectrum (FFT); and (4) Time–frequency spectrogram. The counterfactual exhibits physiologically coherent temporal and spectral characteristics while shifting toward the target prediction, and remains broadly consistent with patterns observed in the natural signals (test instance and NUN).}
	\label{FIG1}
\end{figure*}

Across three PPG datasets, EvoMorph generates novel (Temporal Sparsity=1) yet minimally edited CFE that are structurally and morphologically close to the original test instance (see the example evaluation in Figure \ref{FIG1}) , as reflected by the low proximity and Frequency sparsity. These results indicate that the proposed edit operators and the morphology loss constrain the search to regions preserving the morphology of the data manifold. Achieving high Validity scores, the EvoMorph counterfactuals are also realistic and lie within the desired target output range. Plausibility scores are high and competitive with or superior to those of NUNs—which serve as a conservative plausibility upper bound because they are real samples drawn from the data set by design. The strong plausibility of EvoMorph CFE, particularly for respiratory rate and oxygen saturation, indicates that the constructed time-series reside in label-consistent neighborhoods. A key advantage of EvoMorph over NUN-based surrogates is its ability to generate a large number of distinct counterfactuals per test instance (see Figure \ref{diversityfig}): the diversity counts demonstrate that the evolutionary search explores a rich set of alternative trajectories compatible with both the model and the physiological constraints, which is essential to reveal multiple plausible “what-if” pathways, allowing intuitive reasoning. Figure \ref{divproxval} visualizes the joint relationship between per-instance diversity, waveform sharpness (maximum-gradient), and DTW proximity. Across datasets, the majority of generated CFE cluster at low maximum-gradient while remaining close in DTW, indicating that EvoMorph can explore a large space of distinct counterfactual trajectories without systematically inducing sharp, artifact-like transitions. A smaller subset exhibits elevated maximum-gradient, especially at higher diversity, suggesting occasional over-aggressive edits.

% DIVERSITY PROXIMITY MAX GRADIENT GLOBAL FIGURE PER DATASET

At the same time, the Validity and Maximum Gradient metrics highlight the expected trade-off with in-manifold neighbors: NUNs achieve higher Validity and smoother profiles in two datasets because they reuse unedited training examples, whereas EvoMorph constructs new trajectories and occasionally sacrifices some Validity and smoothness to maintain frequency sparsity and diversity. Importantly, the Maximum Gradient values for  the majority of EvoMorph CFE remain well below those associated with spike-like artifacts (see Figure \ref{divproxval}), confirming that overall signal smoothness and physiological plausibility are preserved. Overall, the strengths of the EvoMorph framework lie in its explicit incorporation of interpretable descriptors, its ability to generate various counterfactuals, and its compatibility with black-box regression models. The morphology-based objective, maximum-gradient constraint, and coherent edit operators together restrict the search to clinically plausible trajectories, resulting in counterfactuals that maintain physiologic coherence across datasets.

In addition to generating CFE, the uncertainty case study on the BIDMCHR dataset shows that EvoMorph counterfactuals retain strong fidelity to the underlying data manifold, as they reproduce uncertainty patterns that closely track those obtained from a bootstrap-based deep ensemble. Here, CFE-based uncertainty is computed as the dispersion of $\hat{f}{(x')}$ over the diverse counterfactual set generated for each test instance, while bootstrap uncertainty is computed as the dispersion of predictions from $B$ models trained on bootstrap-resampled training sets. Across most heart-rate ranges (75–105 bpm), both approaches yield narrow prediction intervals, moderate KDE-based negative log-likelihoods, and low-to-moderate counterfactual variance, indicating that the model is locally well constrained by the data and behaves in a stable, near-interpolatory regime. In contrast, both bootstrap intervals and CFE-induced prediction distributions exhibit systematically increased width, elevated KDE NLL, and higher CFE variance in the 65–75 bpm and 105–115 bpm ranges, consistent with reduced data support and extrapolation behaviour in these regions. Overall, EvoMorph-derived uncertainty patterns closely track uncertainty obtained from a bootstrap ensemble and co-vary with KDE NLL, consistently flagging label ranges where data support is sparse. This agreement suggests that the counterfactual sets generated for each test instance behave like realistic local probes of the data distribution: they explore plausible physiological variations around a given trajectory and reveal whether the prediction is stable to such variations or highly sensitive. This is important because it links counterfactual generation to reliability assessment: beyond offering “what-if” explanations, the same counterfactual neighborhood can identify inputs where the model is likely operating outside well-supported regions. Importantly, counterfactual dispersion is not a calibrated posterior uncertainty; rather, its agreement with an independently obtained ensemble estimate and with density cues provides external validity that it functions as a meaningful heuristic indicator of local epistemic uncertainty. This capability suggests a practical path toward uncertainty-aware, instance-level auditing of TSER models.

% HYPERPARAMATERS TABLE
\begin{table*}[!htbp]
%\centering
\caption{Summary of hyperparameters used for multi-objective optimization, evaluation tolerance, and morphology objective configuration. These parameters kept fixed across datasets.}
\label{tab:hyperparams}
%\begin{tabular*}{l l c c}
\begin{tabular*}{\textwidth}{@{\extracolsep{\fill}} l l c c}
\toprule
\textbf{Group} & \textbf{Parameter} & \textbf{Symbol} & \textbf{Value / Setting} \\
\midrule

% ---------------- NSGA-III ----------------
\multirow{5}{*}{NSGA-III optimization} 
& Population size & $P$ & 100 \\
& Number of generations & $G$ & 50 \\
& Crossover probability & $p_{\text{crossover}}$ & 0.6 \\
& Mutation probability & $p_{\text{mutation}}$ & 0.5 \\
& Initialization window range & [$\gamma, 0.5T$] & [50,\,2000] \\

\midrule

% ---------------- Validity tolerance ----------------
\multirow{1}{*}{Validity and Plausibility tolerance} 
& Regression tolerance interval  & $\delta$  & 5 \\
\midrule
% ---------------- Manifold configuration ----------------
\multirow{6}{*}{Morphology objective configuration}
&   
&   
& \multirow{5}{*}{

\begin{tabular}{cc}
\textbf{Mode} & \textbf{Weight} \\
preserve & 1.0 \\
preserve & 1.0 \\
preserve & 1.0 \\
preserve & 0.5 \\
preserve & 0.5 \\
\end{tabular}
} \\
& Amplitude & $\ell_{\text{amp}}, w_{\text{amp}}$ \\
& Dominant frequency & $\ell_{\text{freq}}, w_{\text{freq}}$ & \\
& Plateau length & $\ell_{\text{plat}}, w_{\text{plat}}$ & \\
& Trend & $\ell_{\text{trend}}, w_{\text{trend}}$ & \\
& Maximum gradient & $\ell_{\text{grad}}, w_{\text{grad}}$ & \\
% ---------------- Edit operators ----------------
\midrule
\multirow{3}{*}{Edit operators}
& Blend factor size & $\beta$ & 1.0 \\
& Phase shift & $\eta$ & 0.5 \\
& Scale  & $\nu$ & 0.5 \\

\bottomrule
\end{tabular*}
\label{hyperparametertable}
\end{table*}

%CONTRAST TO PRIOR WORK
Most prior counterfactual methods for time series are formulated for classification and define validity as a label flip, typically optimized under notions of minimality and proximity that operate in the raw time domain (e.g., TSEvo \cite{Hllig2022}, CoMTE \cite{Ates2021}, Glacier \cite{Wang2024}, and Native-Guide/NUN-guided approaches \cite{Wachter2017}). In these settings, “minimality” is commonly quantified by $L_p$ distances, or local segment perturbations/warps—mechanisms that can preserve numerical closeness without considering waveform morphology. Other lines of work target forecasting rather than extrinsic regression (e.g., ForecastCF \cite{Wang2023}), where the counterfactual goal is defined over future trajectories instead of a user-specified continuous target interval. General recourse-style methods such as FACE \cite{Poyiadzi2020} and MACE \cite{Karimi2019} are not tailored to morphology constraints and therefore do not directly enforce waveform-level plausibility. In contrast, EvoMorph targets TSER, where validity is reaching a user-specified interval of a continuous variable, and where minimality is defined as coherent editing: counterfactuals are minimally modified in a way that preserves morphology and signal plausibility, rather than merely minimizing raw-domain perturbation. Consequently, direct quantitative comparison with prior methods is not possible because both the validity criterion (interval attainment vs. label flip) and the minimality notion (morphology-preserving coherent edits vs. raw-domain proximity/masking) differ. We therefore benchmark against NUNs as a conservative on-manifold baseline and evaluate validity, plausibility, morphology preservation, and diversity explicitly in the regression setting.
 
Practically, EvoMorph enables physiology-constrained “what-if” analysis for TSER models: clinicians and researchers can generate multiple target-controlled waveforms for a given signal and inspect how specific and realistic morphological changes would shift the predicted outcome. Because the framework returns a diverse set of counterfactual trajectories per instance, it supports both intra-individual analysis (how plausible variations around the same patient signal affect the estimate) and inter-individual analysis (how different, still-plausible trajectories can yield similar outcomes), providing example-based understanding. Beyond “what-if” exploration, EvoMorph can be used as a controlled explanatory probe by specifying desired descriptor values (e.g., amplitude, dominant frequency, maximum gradient, trend/plateau statistics) and generating coherent waveforms that satisfy these constraints. Comparing the model’s output in such controlled edits helps identify which physiological descriptor drives the model predictions. This is particularly valuable in modalities or tasks where the main descriptors are not obvious a priori (unlike, e.g., dominant frequency for heart/respiratory rate), and can reveal unexpected model sensitivities. Finally, prediction dispersion over the diverse counterfactuals provides an actionable stability insight for flagging cases where the model appears locally brittle, complementing classical uncertainty estimates without modifying the underlying black-box regressor.

EvoMorph has several limitations. First, its multi-objective search introduces inherent trade-offs by design: improving target attainment or smoothness can reduce morphological similarity and diversity. Second, the morphology-aware constraint is only as expressive as the chosen descriptor set; if relevant physiological structure is not captured by these descriptors, the method may still permit edits that appear plausible under the objective while missing clinically important attributes, and extending to other modalities (e.g., ECG or multi-channel signals) will require re-specifying descriptors and operators. Third, counterfactual quality and diversity are sensitive to practical choices—NSGA-III hyperparameters, stopping criteria, and the size/density of the reference set—which may vary across datasets and can limit performance in low-density regions. Finally, the uncertainty signal derived from dispersion over generated CFE is a heuristic, not a calibrated posterior estimate; its magnitude depends on data density and model smoothness and should be interpreted as an instance-level stability indicator rather than a probability of error, motivating deeper evaluation across model classes and calibration settings. 

Future work should therefore focus on extending the descriptor space, improving optimization robustness, and validating uncertainty signals across models and tasks.

\section{Conclusion}
%NEW
This work demonstrates that physiologically grounded CFE can be generated for TSER use cases by constraining the waveforms to a morphology-based representation and using coherent edit operators to maintain signal plausibility. EvoMorph constructs controlled, minimally altered, and realistic alternatives that support "what-if" analysis for continuous waveforms and enable model auditing by revealing how predictions change under realistic morphological variations. In addition, dispersion over EvoMorph CFE provides a practical means to probe model stability, offering a complementary perspective to classical uncertainty estimation.
Future work should extend EvoMorph to additional modalities (e.g., ECG and multi-channel PPG) and evaluate clinical utility via clinician-facing studies of interpretability and decision-making. Methodologically, extending the search efficiency (e.g., warm starts or surrogate-guided evolution) and refining objective calibration are important next steps to improve the capabilities of the proposed method.

%\section{a}
%The authors declare that they have no known competing financial interests or personal relationships that could have appeared to influence the work reported in this paper.

\printcredits

\appendix

\section{Appendix A. Hyperparameters of CFE Generation}\label{appendixhype}
We chose NSGA-III parameters (population size, number of generations, and crossover/mutation probabilities) based on preliminary experiments to balance convergence and computational cost and keep them fixed across all datasets. We set the tolerance $\delta$ for Validity to reflect clinically meaningful deviations for each target (heart rate, respiratory rate, oxygen saturation), rather than purely statistical variation. To retain consistency, the same $\delta$ tolerance is applied to Plausibility metric. In the morphology configuration, we set the amplitude, dominant frequency, maximum gradient, plateau length, and trend as “preserve” properties with moderate to high weights to anchor the core morphology. Table \ref{hyperparametertable} demonstrates the hyperparameters used during CFE generation.

\section{Appendix B. Case Study Results- Epistemic Uncertainty}\label{appbepistemic}
Table \ref{tab:epistemictable} summarizes epistemic uncertainty statistics stratified by target-value bins. Sample counts are highest in the mid-range bins [75,85) and [85,95) (763 and 733 samples, respectively) and decrease toward the tails, particularly in the uppermost bin [125,135) (47 samples). For most target ranges (75–105 bpm), bootstrap and counterfactual intervals remain relatively narrow, mean KDE NLL is moderate, and counterfactual variance is low to moderate. In contrast, two regions stand out. First, in the 65–75 bin, both interval widths and counterfactual variance increase, together with higher NLL, suggesting reduced data support and increased model uncertainty at lower heart rates. Second, in the 105–115 bin, bootstrap CI width, KDE NLL, and counterfactual variance all peak sharply, indicating a clearly high-uncertainty regime with sparse data and unstable predictions.

\begin{table*}[!htbp]
\centering
%\resizebox{\columnwidth}{!}{%
\begin{tabular}{l|cccccccc}
\toprule
\textbf{Metric | Label Range} 
& \textbf{[55,65)} 
& \textbf{[65,75)} 
& \textbf{[75,85)} 
& \textbf{[85,95)} 
& \textbf{[95,105)} 
& \textbf{[105,115)} 
& \textbf{[115,125)} 
& \textbf{[125,135)} \\
\midrule
\textbf{Samples ($n$)} 
& 125 & 191 & 763 & 733 & 284 & 150 & 103 & 47 \\

\textbf{Mean $y_{true}$} 
& 62.78 & 72.14 & 80.81 & 89.64 & 97.00 & 110.46 & 123.04 & 126.22 \\

\textbf{BTS CI Width} 
& 2.82 & 3.36 & 1.77 & 1.72 & 2.48 & 13.17 & 2.13 & 2.46 \\

\textbf{CFE CI Width} 
& 2.53 & 3.78 & 2.50 & 2.88 & 3.14 & 4.91 & 4.06 & 3.88 \\

\textbf{KDE NLL (mean)} 
& 8.49 & 12.46 & 7.06 & 8.37 & 9.27 & 28.61 & 11.32 & 10.29 \\

\textbf{CFE Variance (mean)} 
& 8.94 & 22.75 & 11.34 & 13.17 & 16.43 & 48.49 & 24.23 & 21.76 \\
\bottomrule
\end{tabular}
%}
\caption{Epistemic uncertainty results across target-value bins in the BIDMC32HR dataset. Mean values and CI widths rounded to two decimals.}
\label{tab:epistemictable}
\end{table*}

\section{Appendix C. Inception Architecture and Training Details} \label{appendixmlarc}
The regressor used in this study is a 1D Inception Network deep convolutional neural network, consisting of six inception modules, each contains a 1×1 Conv1D (64 filters) bottleneck, followed by six parallel Conv1D branches with 32 filters and varying kernel sizes (64,32,32,16,32,32), MaxPooling1D and Conv1D branch. The branch outputs are concatenated, batch-normalized, and passed through ReLU activation. Every third module uses a residual shortcut implemented as a 1×1 Conv1D + BatchNorm added to the module output, followed by ReLU. A GlobalAveragePooling1D layer aggregates final temporal features, and a final linear dense layer produces the scalar regression output. We train the network end-to-end with the Huber loss (delta=3.0), Adam optimizer (learning rate=0.001) and MAE as metric, with a piecewise learning rate schedule that applies multiplicative decay at later epochs. We train the network for 40 epochs with the batch size of 32.

\section{Appendix D. Multi-Objective Optimization Parameters}\label{objmoe}
\begin{figure*}[!htbp]
	\centering
	\includegraphics[width=\textwidth]{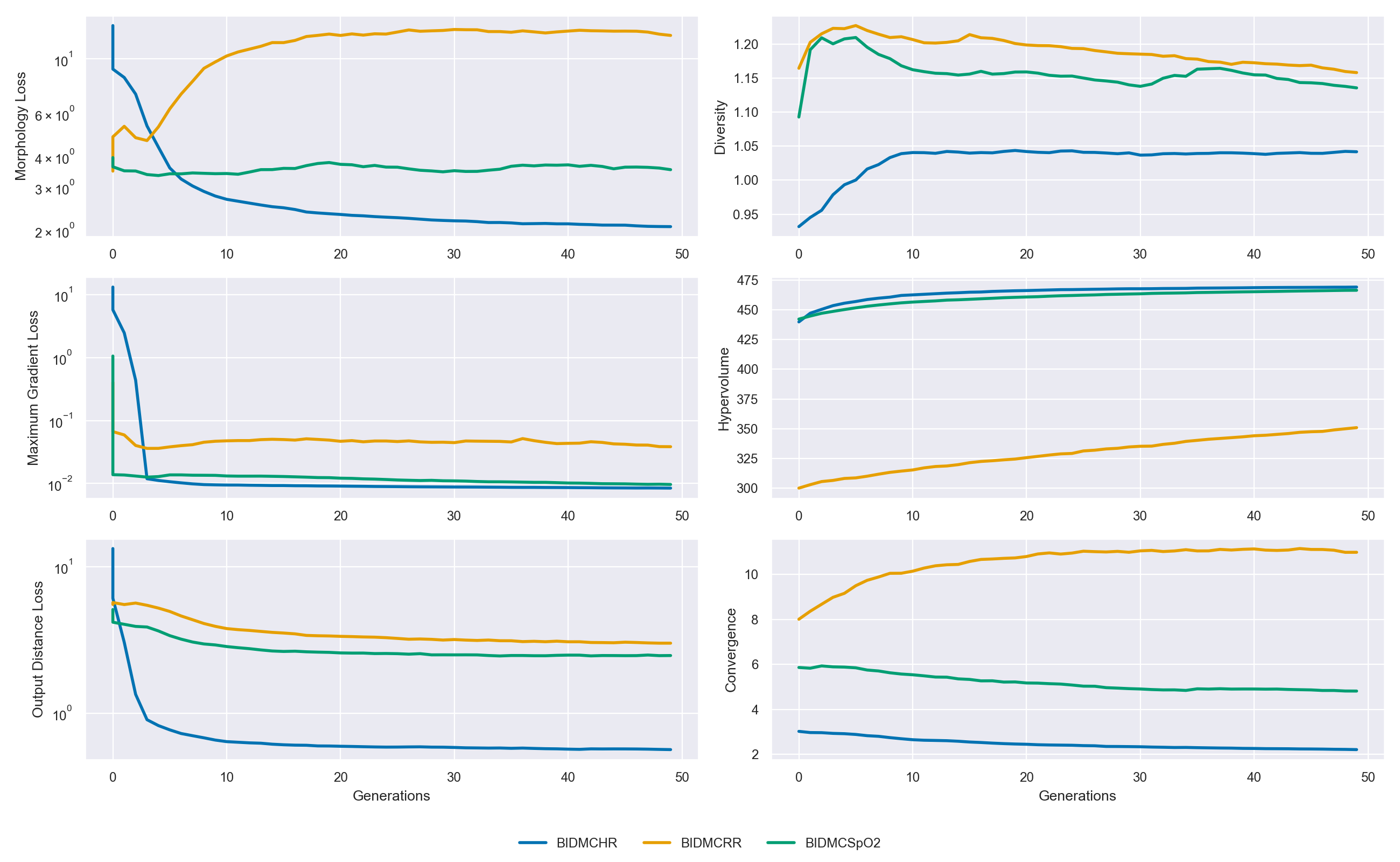}
	\caption{Aggregated evolution of NSGA-III objectives and algorithmic diagnostics across all test instances. Left column: median trajectories for the three optimization objectives, Morphology, Maximum Gradient, and Output Distance losses, over 50 generations. Right column: NSGA-III performance metrics, including population diversity, hypervolume, and convergence.}
	\label{nsgahypesandconv}
\end{figure*}

Figure \ref{nsgahypesandconv} summarizes the behavior of EvoMorph’s NSGA-III optimization across datasets, aggregated over all test instances and 50 generations. The left panels show the evolution of the three objectives. For $\mathcal{O}_{morph}$, the median objective value decreases sharply during the first 5–10 generations for all datasets and then levels off, with BIDMCHR reaching the lowest final values and RR/SpO2 stabilizing at slightly higher levels. This indicates that counterfactuals increasingly align with the modality-specific property constraints and that most improvement occurs early in the evolutionary run. Additionally, the achievable level of property fidelity differs across datasets, as this objective plateaus at different levels.

$\mathcal{O}_{maxgrad}$ converges even more rapidly, dropping by roughly one to two orders of magnitude within a few generations and remaining stable thereafter, suggesting that gradient-based smoothness constraints are quickly satisfied and maintained. $\mathcal{O}_{out}$ shows a similar pattern: initial large output deviations are reduced substantially within the first 5–10 generations, after which counterfactual predictions remain close to the original regression outputs. Together, these trajectories indicate that the populations used in our experiments correspond to near-converged trade-offs between property fidelity, local smoothness, and output consistency.

The right panels report standard NSGA-III diagnostics. Diversity remains strictly positive for all datasets, with BIDMCRR and SpO2 exhibiting higher steady-state diversity than HR, indicating that the algorithm maintains a heterogeneous set of counterfactual solutions rather than collapsing to a single mode. Hypervolume increases monotonically and saturates after ~10–20 generations, reflecting continuous improvement and subsequent stabilization of the Pareto front in objective space. The convergence metric shows decreasing or plateauing trends, with more dynamic behaviour for BIDMCRR, consistent with its more challenging objective landscape. Overall, these diagnostics confirm that NSGA-III performs a stable, well-behaved multi-objective search, and that the counterfactual ensembles analyzed in the main experiments are drawn from populations that have reached a mature Pareto front with adequate diversity.

%% Loading bibliography style file
%\bibliographystyle{model1-num-names}
\bibliographystyle{cas-model2-names}

% Loading bibliography database
\bibliography{cas-refs}

\end{document}